\documentclass[runningheads]{llncs}
\usepackage{graphicx}
\usepackage{here}
\usepackage{comment}
\usepackage{amsmath}
\usepackage{amssymb}
\usepackage{amsfonts}
\usepackage{amsmath}
\usepackage{color}
\usepackage[normalem]{ulem}

\begin{document}
\title{Learning State Transition Rules from Hidden Layers of Restricted Boltzmann Machines}
\author{Koji Watanabe\inst{1,2} \and Katsumi Inoue\inst{1,2}}
\authorrunning{K.\ Watanabe and K.\ Inoue}

\institute{The Graduate University for Advanced Studies, SOKENDAI, Tokyo, Japan \and National Institute of Informatics, Tokyo, Japan \\
\email{\{kojiwatanabe,inoue\}@nii.ac.jp}}
\maketitle

\begin{abstract}
Understanding the dynamics of a system is important in many scientific and engineering domains. This problem can be approached by learning state transition rules from observations using machine learning techniques. Such observed time-series data often consist of sequences of many continuous variables with noise and ambiguity, but we often need rules of dynamics that can be modeled with a few essential variables. In this work, we propose a method for extracting a small number of essential hidden variables from high-dimensional time-series data and for learning state transition rules between these hidden variables. The proposed method is based on the Restricted Boltzmann Machine (RBM), which treats observable data in the visible layer and latent features in the hidden layer. However, real-world data, such as video and audio, include both discrete and continuous variables, and these variables have temporal relationships. Therefore, we propose Recurrent Temporal GaussianBernoulli Restricted Boltzmann Machine (RTGB-RBM), which combines Gaussian-Bernoulli Restricted Boltzmann Machine (GB-RBM) to handle continuous visible variables, and Recurrent Temporal Restricted Boltzmann Machine (RT-RBM) to capture time dependence between discrete hidden variables. We also propose a rule-based method that extracts essential information as hidden variables and represents state transition rules in interpretable form. We conduct experiments on Bouncing Ball and Moving MNIST datasets to evaluate our proposed method. Experimental results show that our method can learn the dynamics of those physical systems as state transition rules between hidden variables and can predict unobserved future states from observed state transitions.
\end{abstract}

\section{Introduction}
Learning the dynamics of a system from data is important in many scientific and engineering problems. 
We express rules of the dynamic by various symbolic forms such as equations, programs, and logic to understand the dynamics because they are usually simple yet explicitly interpretable and general. With advances in computing power and Internet technologies, the data we handle, such as video and audio, is becoming increasingly massive. Moreover, the data observed from dynamics often consist of many continuous sequences of variables and contain noise and ambiguity. Therefore, finding rules from such large data is becoming more difficult. 

To address this problem, several methods have been proposed to learn rules from large data by combining symbolic regression with deep learning \cite{udrescu2020ai,cranmer2020discovering}. 
These methods make it possible to model the dynamics as equations and predict the future and past. While some dynamics can be expressed in quantitative relationships, such as classical mechanics and electromagnetism, some dynamics are expressed by state transition rules, such as \textit{Boolean networks} (BNs) \cite{kauffman1993origins} and \textit{Cellular Automata} (CA) \cite{wolfram2018cellular}.
\textit{Learning from interpretation transition} (LFIT) \cite{inoue2014learning} is an unsupervised learning algorithm that learns the rules of the dynamics from state transitions. The LFIT framework learns state transition rules as a \textit{normal logic program} (NLP). Some methods have been proposed combining LFIT with \textit{neural networks} (NNs) to make them more robust to noisy data and continuous variables. For example, NN-LFIT \cite{gentet@2016} extracts propositional logic rules from trained NNs, D-LFIT \cite{gao2022learning} translates logic programs into embeddings, infers logical values through differentiable semantics of the logic programs, and searches for embeddings using optimization methods and NNs. 
However, while these LFIT-based methods can learn rules between observable variables, they cannot learn rules for unobservable variables. It is not necessary to use all observable variables to explain dynamics; often, only a few key factors and their relationships can explain an essential part of dynamics. Such factors are not always contained in the observable data and are sometimes unobservable. Also, when the original dynamics are composed of a large number of observable variables, the rules describing their relationships may suffer from exponential combination problems.
Therefore, it is important to extract a small number of essential factors from observable variables as hidden variables and express their relationships as rules.

Several approaches have been proposed for learning symbolic representations using hidden variables by \textit{restricted Boltzmann Machine} (RBM) \cite{rabiner1986introduction,hinton2002training}.
Propositional formulas are expressed in RBM, and symbolic knowledge is learned as maximum likelihood estimation of unsupervised energy functions of RBM \cite{tran2017propositional}. \textit{Logical Boltzmann Machine} (LBM) \cite{tran@2021} is a \textit{neuro-symbolic} system that converts any propositional formula described in DNF into RBM and uses RBM to achieve efficient reasoning. LBM can be used to show the equivalence between minimizing the energy of RBM and the satisfiability of Boolean formulae. 
However, these approaches cannot learn hidden representations from raw data such as images.
Therefore, several other neural network-based methods have been proposed for learning meaningful hidden representations from raw data.
Some are based on other generative models such as VAE \cite{kingma2013auto}. For example, $\beta$-VAE \cite{higgins2016beta} learns each dimension of the hidden variable to have as much disentangled meaning as possible, and joint-VAE \cite{dupont2018learning} controls the output by treating the hidden variable as a categorical condition \cite{davi2018d@worldmodels}. While these approaches can extract meaningful hidden representations from data, they cannot handle hidden representations in interpretable forms such as symbolic representations or rules.

In this study, we propose a method that can learn interpretable symbolic representations from observation data. We aim to extract a few essential hidden variables sufficient to predict the dynamics and then learn state transition rules between these hidden variables.
The proposed method is based on RBM, which treats observable data as visible variables and latent features as hidden variables.
Real-world data, such as video and audio, include both discrete and continuous values, which have temporal relationships. Therefore, we propose \textit{recurrent temporal Gaussian-Bernoulli restricted Boltzmann Machine} (RTGB-RBM) \cite{sutskever2008recurrent,mittelman2014structured}, which combines \textit{Gaussian-Bernoulli restricted Boltzmann Machine} (GB-RBM) \cite{hinton2006reducing} to handle continuous visible variables, and \textit{recurrent temporal restricted Boltzmann Machine} (RT-RBM) to capture time dependence between discrete latent variables.
We conduct experiments on a Bouncing Ball dataset generated by a \textit{neural physics engine} (NPE) \cite{chang2016compositional}, and \textit{Moving MNIST} \cite{srivastava2015unsupervised}. Experimental results show that our method can learn the dynamics of those physical systems as state transition rules between latent variables and predict unobserved future states from observed state transitions. 

This paper is structured as follows. We will first cover some necessary background of our method in Section \ref{sec2}. Next, we will present our method in Section \ref{sec3}, then show our experimental results in Section \ref{sec4}. Finally, we will summarize our work and discuss some possible further research in Section \ref{sec5}.

\vspace{-0.1cm}
\section{Background} \label{sec2}
\subsection{Gaussian-Bernoulli Restricted Boltzmann Machine} \label{sec21}
\begin{figure}[H]
  \vspace{-0.3cm}
  \centering
  \includegraphics[width=4cm]{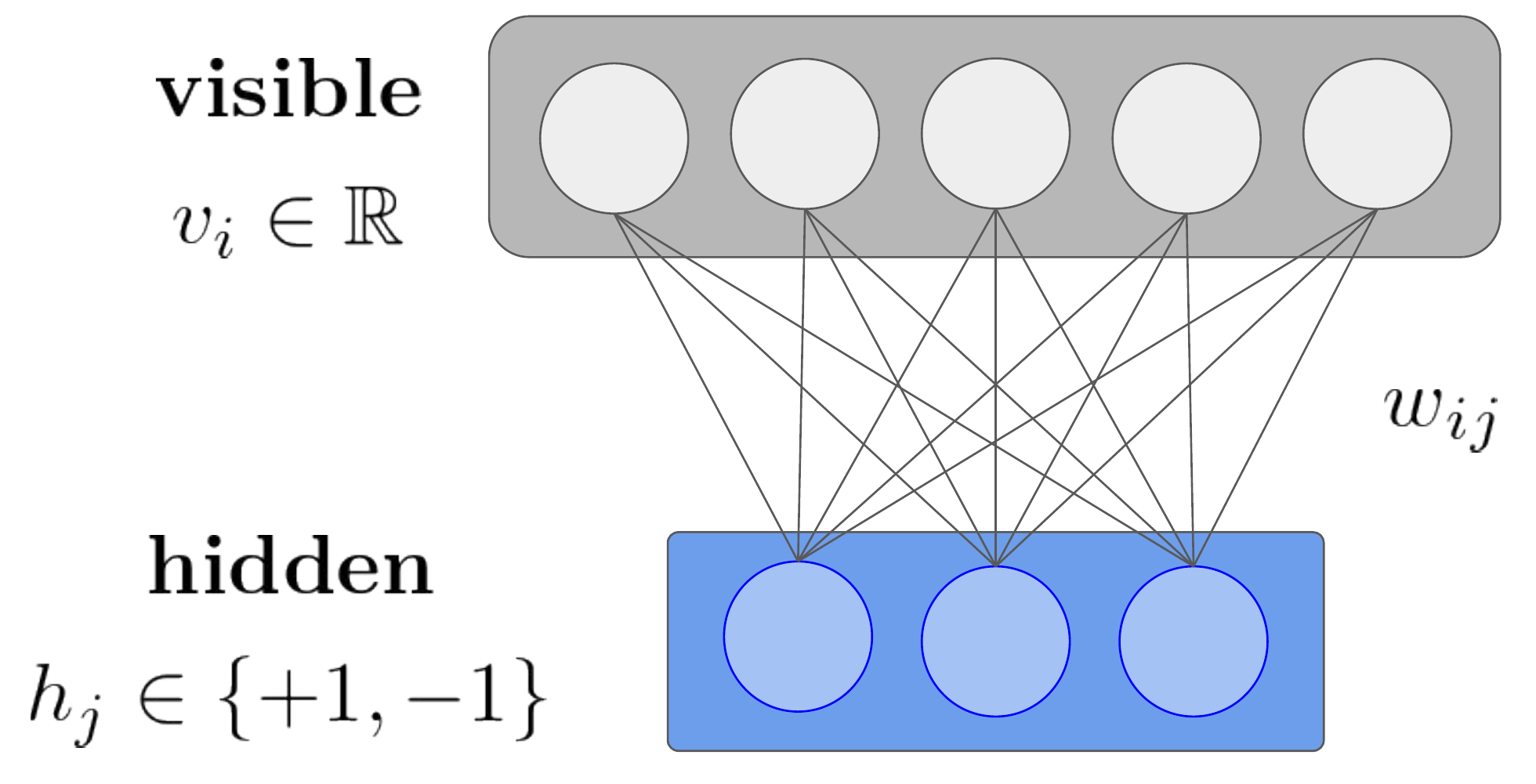}
  \vspace{-5pt}
  \caption{Graphical representation of GB-RBM}
  \label{fig_gbrbm}
\end{figure}

A Gaussian-Bernoulli restricted Boltzmann Machine (GB-RBM) \cite{hinton2006reducing} is defined on a complete bipartite graph as shown in Figure \ref{fig_gbrbm}. 
The upper layer is the visible layer $V$ consisting of only visible variables, and the lower layer is the hidden layer $H$ consisting of only hidden variables, where $V$ and $H$ are the index of visible and hidden variables, respectively.
\vspace{-0.4cm}
\vspace{-0.7cm}
\noindent
$\mathbf{v} = \{v_i \in \mathbb{R} \ | \ i \in V \}$ represents real variables directly associated with the input-output data, and $\mathbf{h} = \{h_j \in \{+1, -1\} \ | \ j \in H \}$ represents hidden variables of the system that is not directly associated with the input-output data and is a discrete variable taking binary values. $\mathbf{s} = \{s_i \ | \ i \in V\}$ is the parameter associated with the variance of the visible variables.
The energy function of the GB-RBM is defined as 
\vspace{-0.2cm}
\begin{eqnarray}  \label{rgbm_energy}
  E_{\theta}(\mathbf{v},\mathbf{h}) &=&
  \sum_{\ i \in V} \frac{(v_i - b_i)^2}{2 s_i^2}
  + \sum_{\ i \in V} \sum_{\ j \in H} \frac{w_{ij}}{s_i^2}v_i h_j
  + \sum_{\ j \in H} c_j h_j
\end{eqnarray}
\noindent
Here, $\mathbf{b} = \{b_i \ | \ i \in V \}$ and $\mathbf{c} = \{c_j \ | \ j \in H\}$ are the bias parameters for the visible and hidden variables respectively.
$\mathbf{w} = \{w_{ij} \ | \ i \in V, j \in H\}$ is the set of parameters between the visible and hidden variables. $\mathbf{s} = \{s_i \ | \ i \in V \}$ is a variance of the visible variables. These model parameters are collectively denoted by $\theta = \{\mathbf{W},\mathbf{b}, \mathbf{c} \} $. 
Using the energy function in (\ref{rgbm_energy}), the joint probability distribution of $\mathbf{v},\mathbf{h}$ is defined as
\begin{eqnarray}
\begin{aligned}
  & P_{\theta}(\mathbf{v},\mathbf{h}) = \frac{1}{Z(\theta)} \mbox{exp}(-E_\theta\left(\mathbf{v},\mathbf{h})\right) \\
  & Z(\theta) = \int_{-\infty}^{+\infty} \sum_{\mathbf{h}} \mbox{exp}\left(-E_\theta(\mathbf{v},\mathbf{h})\right) d\mathbf{v}
\end{aligned}
\end{eqnarray}
\noindent
$Z_\theta$ is a \textit{partition function}, $\int_{-\infty}^{+\infty}...d\mathbf{v}$ represents the multiple integral with respect to $\mathbf{v}$, $\sum_{\mathbf{h}}$ represents the multiple sum over all possible combinations of $\mathbf{h}$.
The conditional probability distributions of $\mathbf{v}$ and $\mathbf{h}$, given $\mathbf{h}$ and $\mathbf{v}$, are respectively 
\begin{eqnarray}
\vspace{-0.2cm}
P_{\theta}(v_i = v \ |\ \mathbf{h}) &=& \mathcal{N} (v_{i} \ | \ b_i + \sum_j w_{ji} h_{j}, s_i^2) \label{gbm_pv}\\
P_{\theta}(h_j = 1 \ |\ \mathbf{v}) &=& \frac{\exp \left( c_j +\sum_i w_{ij}v_i \right)} {2 \mbox{cosh} \left( c_j + \sum_i w_{ij}v_j\right)} \label{gbm_ph}
\vspace{-0.2cm}
\end{eqnarray}
By (\ref{gbm_pv}), given $\mathbf{h}$, the probability of $v_{i} = v$ is calculated, and we can sample $v_{i}$ from the probability. By (\ref{gbm_ph}), given $\mathbf{v}$, the probability of $h_{j}=1$ is calculated, and we can sample $h_{j}$ from the probability
\subsection{Recurrent Temporal Restricted Boltzmann Machine} \label{sec22}
\begin{figure}[H]
  \vspace{-0.7cm}
  \centering
  \includegraphics[width=5cm]{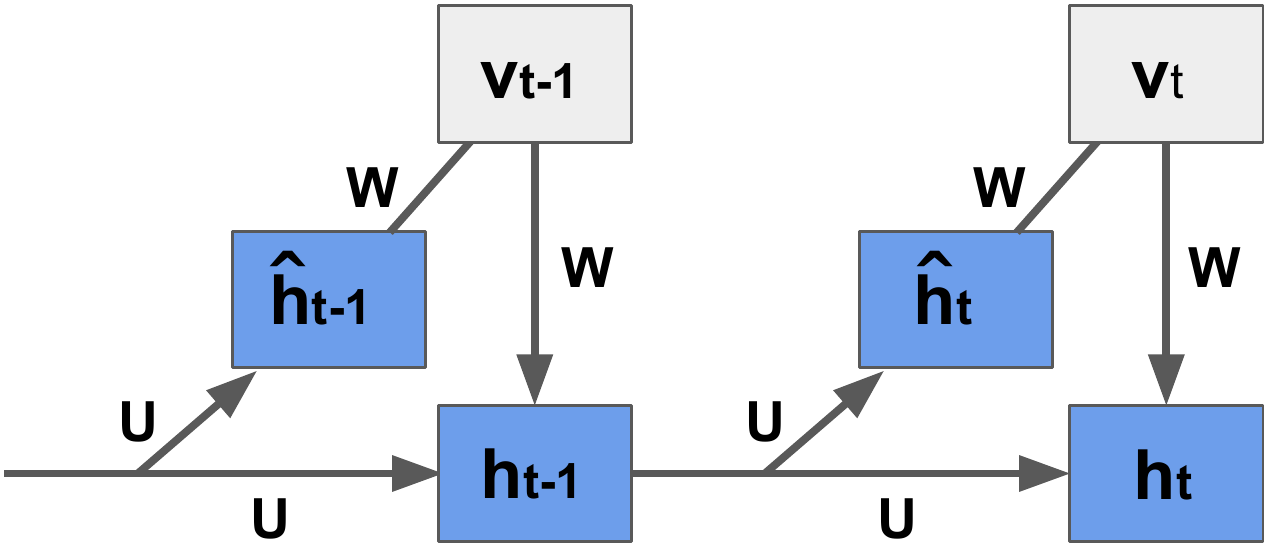}
  \vspace{-5pt}
  \caption{Graphical representation of RT-RBM}
  \label{fig_rtrbm}
\end{figure}

A recurrent temporal restricted Boltzmann Machine (RT-RBM) is an extension of RBM \cite{sutskever2008recurrent,mittelman2014structured} and is suitable for handling time series data. 
The RT-RBM has a structure with connections from the set of hidden variables from the past $k$ frames $\{ \mathbf{h}_{t-k},\mathbf{h}_{t-k+1},...,\mathbf{h}_{t-1}\}$ to the current visible variables $\mathbf{v}_t$ and hidden variables $\mathbf{h}_t$.
This paper assumes that the state at time $t$ depends only on one previous time state $t-1$ and fix $k=1$.
The RT-RBM for $k=1$ is shown in Figure \ref{fig_rtrbm}.
In RT-RBM, in addition to parameters $\mathbf{W}$, $\mathbf{b}$ and $\mathbf{c}$, which are defined in Sec \ref{sec21}, we newly define $\mathbf{U} = \{u_{jj'} \ | \ j,j' \in H\}$, which is the set of parameters between the hidden variables at time $t$ and $t-1$. These model parameters are collectively denoted by $\theta = \{\mathbf{W}, \mathbf{U}, \mathbf{b}, \mathbf{c}\}$.
Then, the expected value of the hidden vector $\hat{\mathbf{h}}_t$ at time t is defined as,
\begin{eqnarray} \label{h_hat}
\hat{\mathbf{h}}_t = 
\begin{cases}
\sigma \left(\mathbf{W}\mathbf{v}_t + \mathbf{c} + \mathbf{U}\hat{\mathbf{h}}_{t-1}\right),  \mbox{ if }t > 1 \\
\sigma \left(\mathbf{W}\mathbf{v}_t + \mathbf{c} \right), \ \ \ \ \  \mbox{ if }t = 1 
\end{cases}
\end{eqnarray}
where $\sigma$ is a sigmoid function $\sigma(x) = (1 + \mbox{exp}(-x))^{-1}$.
\noindent
Given $\hat{\mathbf{h}}_{t-1}$, The conditional probability distributions of $v_{t,i}$ and $h_{t,j}$ are inferenced by 
\begin{eqnarray}
P_\theta(v_{t,i}=1 \ | \ \mathbf{h}_{t} , \hat{\mathbf{h}}_{t-1}) &=& \sigma ( \sum_{j \in H} w_{ji} h_{t,j} + b_i ) \label{rtrbm_pv}  \\
P_\theta(h_{t,j}=1 \ | \ \mathbf{v}_{t} , \hat{\mathbf{h}}_{t-1}) &=& \sigma ( \sum_{i \in V} w_{ji} v_{t,i} + c_j + \sum_{j' \in H} u_{jj'} \hat{h}_{t-1,j'}   )  \label{rtrbm_ph}
\end{eqnarray}
By (\ref{rtrbm_pv}), given $\mathbf{h}_t$ and $\mathbf{\hat{h}}_t$, the probability of $v_{t,i}=1$ is calculated, and we can sample $v_{t,i}$ from the probability. By (\ref{rtrbm_ph}), given $\mathbf{v}_t$ and $\mathbf{\hat{h}}_t$, the probability of $h_{t,j}=1$ is calculated, and we can sample $h_{t,j}$ from the probability.

\vspace{-0.8cm}
\section{Learning State Transition Rules from RBM} \label{sec3}
This section proposes a new method called recurrent temporal Gaussian-Bernoulli restricted Boltzmann Machine (RTGB-RBM), which integrates GB-RBM (to handle continuous visible variables) with RT-RBM (to capture time dependencies between discrete hidden variables). The RTGB-RBM takes $\{\mathbf{v}_0,\mathbf{v}_1,...,\mathbf{v}_t \}$ as input and and predicts the future states $\{ \mathbf{v}_{t+1},\mathbf{v}_{t+2},...,\mathbf{v}_{T}\}$. In addition, we extract a set of state transition rules from the trained RTGB-RBM. The rules represent the original dynamics using a few essential hidden variables. State transition rules are described as
\begin{eqnarray} \label{rule_form}
   p_j :: L_{t+1,j} \leftarrow L_{t,1} \wedge L_{t,2} \wedge L_{t,3} \wedge .... \wedge \ L_{t,m} 
\end{eqnarray}
where, $m$ is the number of hidden variables in each hidden layer, $L_{t,j} (1 \leq j \leq m)$ is a literal that represents a hidden variable $h_{t,j}$
or its negation $\neg h_{t,j}$, and $p_j$ is the probability of occurring the $j$-th rule. For example, suppose we get a rule $ 0.8::  h_{t+1,1}\leftarrow h_{t,1} \wedge h_{t,2} \wedge \neg h_{t,3}$.
This rule represents that, if $h_{t,1}=1, h_{t,2}=1, h_{t,3}=0$, then $h_{t+1,1}$ will be $1$ with the probability $0.8$.
\subsection{Recurrent Temporal Gaussian-Bernoulli RBM}
This subsection describes the Recurrent Temporal Gaussian-Bernoulli Restricted Boltzmann Machine (RTGB-RBM).
RTGB-RBM is defined by the set of parameters $\theta = \{\mathbf{W},  \mathbf{U}, \mathbf{b},\mathbf{c}, \mathbf{s}\}$. 
These parameters are introduced in Sec \ref{sec21} and \ref{sec22}.
In RT-RBM, both visible and hidden variables take binary values, while in RTGB-RBM, visible variables take continuous values, and hidden variables take binary values. This difference makes it possible to handle data with a wide range of values, such as images, in the visible layer and to handle their features in the hidden layer. The difference between GB-RBM and RTGB-RBM is that GB-RBM cannot handle sequences, while RTGB-RBM can handle sequences in both visible and hidden layers by defining weights for transitions between hidden layers. By combining RT-RBM and GB-RBM, RTGB-RBM can learn time series data with a wide range of values, such as video and sound.
$\mathbf{v}_t$ and $\mathbf{h}_t$ are inferenced by the following equations,
\begin{eqnarray}
\vspace{-0.2cm}
P_{\theta}(v_{t,i} \ |\  \mathbf{h}_t,\ \hat{\mathbf{h}}_{t-1}) &=& \mathcal{N}(v_{t,i} \ | \ b_{i} + \sum_j w_{ji} h_{t,j}  \ ,s_i^2) \label{p_vt_ht} \label{rtgbrbm_pv} \\
P_{\theta}(h_{t,j}=1 \ |\  \mathbf{v}_t,\ \hat{\mathbf{h}}_{t-1}) &=&  \sigma(\sum_i w_{ji} \frac{v_{t,i}}{s^2} + c_{j} + \sum_{j'}u_{jj'}\hat{h}_{t-1,j'}) \label{rtgbrbm_ph}
\vspace{-0.2cm}
\end{eqnarray}
$\hat{\mathbf{h}}_{t}$ is calculated by (\ref{h_hat}).
By (\ref{rtgbrbm_pv}), given $\mathbf{h}_t$ and $\mathbf{\hat{h}}_t$, the probability of $v_{t,i}$ is calculated, and we can sample $v_{t,i}$ from the probability. By (\ref{rtgbrbm_ph}), given $\mathbf{v}_t$ and $\mathbf{\hat{h}}_t$, the probability of $h_{t,j}=1$ is calculated, and we can sample $h_{t,j}$ from the probability.

\subsection{Training}
We update the parameters of RTGB-RBM so that the likelihood $L$ is maximized.
\begin{eqnarray*}
 L &=& \prod_n^N \prod_t^T P_{\theta}(\mathbf{v}_{t}^{(n)}| \ \mathbf{h}_{t}^{(n)}, \hat{\mathbf{h}}_{t-1}^{(n)}) 
\end{eqnarray*}
The parameter $\mathbf{\theta} = \{\mathbf{W},\mathbf{U}, \mathbf{b}, \mathbf{c}\}$ that maximizes the product of $\mathbf{v}_t$ is estimated by the gradient method $\mathbf{\theta} = \mathbf{\theta} + \frac{\partial log L}{\partial \theta}$.
The gradients of each parameter are calculated as follows
\begin{eqnarray}
\hspace{-1.0cm}
\begin{aligned}
  & \frac{\partial log L}{\partial w_{ij}} = \langle \frac{v_{t,i} \hat{h}_{t,j}}{s_i^2}   \rangle_{data} -  \langle \frac{v_{t,i} \hat{h}_{t,j}}{s_i^2}  \rangle_{model}, \ \ \ \ \ \ \ \
     \frac{\partial log L}{\partial b_{i}} =  \langle v_{t,i} \rangle_{data} - \langle v_{t,i} \rangle_{model} \\
  & \frac{\partial log L}{\partial u_{jj'}} =  \langle\hat{h}_{t-1,j'}\hat{h}_{t,j} \rangle_{data} - \langle\hat{h}_{t-1,j'}\hat{h}_{t,j} \rangle_{model}, \ \  
  \frac{\partial log L}{\partial c_{j}} =  \langle \hat{h}_{t,j} \rangle_{data} - \langle \hat{h}_{t,j} \rangle_{model} 
\end{aligned}
\end{eqnarray}
where, $\langle x \rangle_{data}$ represents the mean of $x$. 
$\langle x \rangle_{model}$ represents the expected value of $x$ , and it can be calculated using Contrastive Divergence (CD) algorithm \cite{hinton2006fast}. 
Learning $w_{ij}$ and $b_i$ increase the accuracy of reconstruction data in the visible layer from the hidden layer, and learning $u_{jj'}$ and $c_{j}$ increases the accuracy of predicting transitions.
\subsection{Extracting Transition Rules}
We extract state transition rules between hidden variables from the conditional probability distribution $P_\theta(\mathbf{h}_t | \mathbf{h}_{t-1})$. 
We cannot easily compute  $P_\theta(\mathbf{h}_t | \mathbf{h}_{t-1})$. Therefore, we approximate $P_\theta(\mathbf{h}_t | \mathbf{h}_{t-1})$ using Gibbs sampling.
Here we define $\mathbf{v}_t(k)$ and $\mathbf{h}_t(k)$ as vectors obtained by repeating Gibbs sampling $k$ times, and assume $\mathbf{h}_{t-1}$ is given.
Then, $P_\theta(\mathbf{h}_t | \mathbf{h}_{t-1})$ is approximated by the following steps \\
$\mbox{     }$ \textbf{Step1:} Generate $\mathbf{h}_t(0)$ randomly. \\
$\mbox{     }$ \textbf{Step2:} Sample $\mathbf{v}_t(1)$ from $P_\theta(\mathbf{v}_t(1) | \mathbf{h}_t(0), \mathbf{h}_{t-1} )$. \\
$\mbox{     }$ \textbf{Step3:} Sample $\mathbf{h}_t(1)$ from $P_\theta(\mathbf{h}_t(1) | \mathbf{v}_t(0), \mathbf{h}_{t-1} )$. \\
$\mbox{     }$ \textbf{Step4:} Repeat \textbf{steps2,3} $k$ times, and get $\mathbf{h}_t(k)$. \\
$\mbox{     }$ \textbf{Step5:} Approximate $P_\theta(\mathbf{h}_t \ | \mathbf{h}_{t-1})$ as $P_\theta(\mathbf{h}_t(k) | \mathbf{h}_{t-1})$. \\
By repeating the above steps, we get $\mathbf{v}_{t}(k)$ and $\mathbf{h}_{t}(k)$ from $\mathbf{v}_{t}(0)$ and $\mathbf{h}_{t}(0)$ as,
\begin{eqnarray*}
\mathbf{h}_t(0) \rightarrow \mathbf{v}_t(0) \rightarrow \mathbf{h}_t(1) \rightarrow \mathbf{v}_t(1)  \rightarrow ... \rightarrow \mathbf{h}_t(k) \rightarrow \mathbf{v}_t(k) 
\end{eqnarray*}
If we take $k$ large enough, we can approximate $P_\theta(\mathbf{h}_t | \mathbf{h}_{t-1})$ well.
The transition rules between the hidden variables are extracted from $P_\theta(\mathbf{h}_t | \mathbf{h}_{t-1})$, by computing the combination of $\mathbf{h}_t$ and $\mathbf{h}_{t-1}$. We determine $\mathbf{h}_t$ from $\mathbf{h}_{t-1}$ using extracted rules and decode $\mathbf{v}_t$ from $\mathbf{h}_t$ by (\ref{p_vt_ht}).

\begin{figure}[H]
  \vspace{-0.6cm}
  \centering
  \includegraphics[width=10cm]{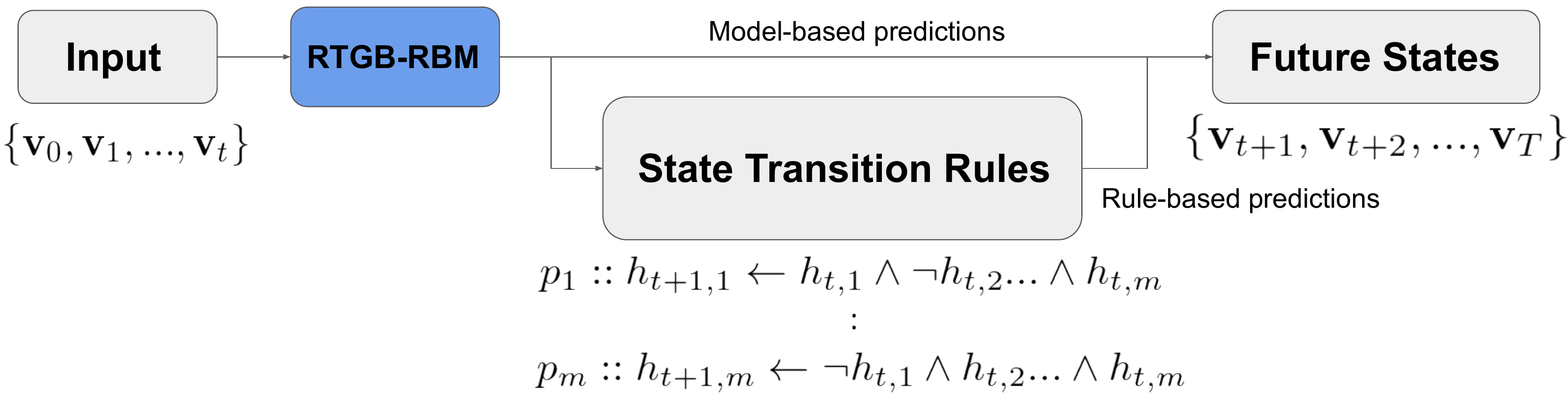}
  \vspace{-0.2cm}
  \caption{Overview of our method. We extract transition rules from approximated conditional
probability distribution $P_\theta(\mathbf{h}_t | \mathbf{h}_{t-1})$ of trained RTGB-RBM.}
  \label{overview_our_method}
\end{figure}

The overview of our method is illustrated in Figure \ref{overview_our_method}. 
We have two types of predictions: model-based predictions and rule-based predictions. Given the observed state sequence $\{ \mathbf{v}_{0},\mathbf{v}_{1},...,\mathbf{v}_{t}\}$ as input, the former predicts future states using RTGB-RBM and the latter predicts future states by interpretable rules expressed in equation (\ref{rule_form}) extracted from the trained RTGB-RBM.
\section{Experiments} \label{sec4}
In this study, we conducted experiments on a Bouncing Ball dataset generated by the neural physics engine (NPE) \cite{chang2016compositional}.
The dataset is a simulation of multiple balls moving around in a two-dimensional space surrounded by walls on all four sides. The number of balls, radius, color, speed, etc., can be changed. 
\subsection{Setting}
We experimented with two videos, one with x1 ball and one with x3 ball, where the pixels are $[0,1]$. 
In both cases, x1 ball and x3 ball, we generated 10000 videos of size 100x100 pixels and duration of 100 time-steps.
First, we train RTGB-RBM to predict the future state of the balls, then extract transition rules between hidden variables.
The weights are updated after computing the gradient on a single sequence. The number of CD iterations during training is set to $20$.
To evaluate the predictions of our model, we used  (\ref{loss}), it indicates loss of the prediction $\{ \hat{\mathbf{v}}_{t=T+1},...,\hat{{v}}_{T'} \}$, given $\{\mathbf{v}_{0},...,{v}_{T}\}$. Here, we set $N=10000$, $T=90$, $T'=100$. $\mathbf{v}_{t}$ is the observed data, $t$, $\hat{\mathbf{v}_t}$ is the prediction at the time $t$.
\begin{eqnarray} \label{loss}
\vspace{-1cm}
\mbox{Loss} = \frac{1}{N}\sum_{n=0}^N \left( \frac{1}{T'-T}\sum_{t=T+1}^{t=T'} \sum_{i \in V} (v_{t,i}^{(n)} - \hat{v}_{t,i}^{(n)})^2 \right)
\end{eqnarray}
\subsection{Training}
\begin{figure}[H]
  \vspace{-0.5cm}
  \begin{center}
    \includegraphics[width=10cm]{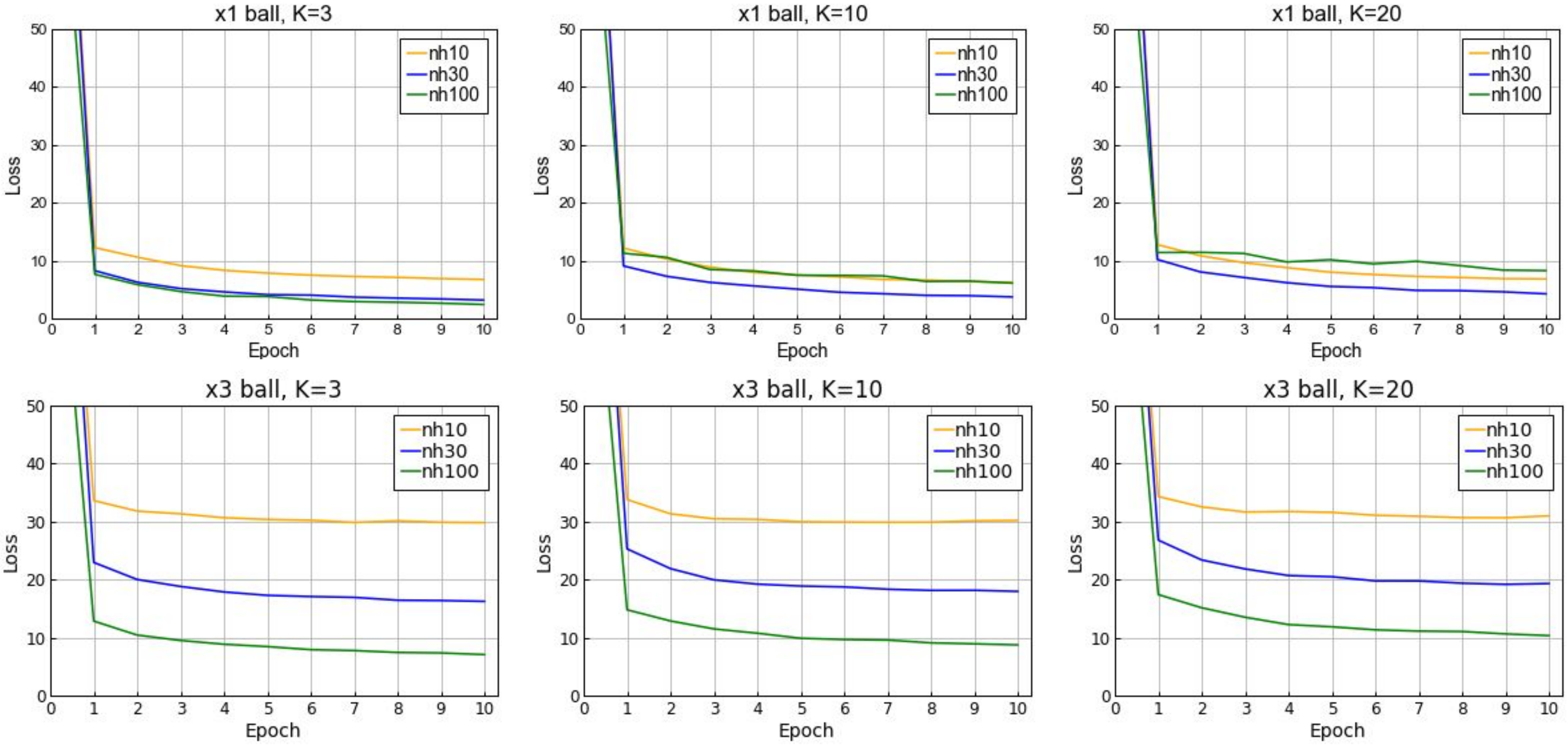}
    \caption{Learning curve of x1 ball case (top) and x3 ball (bottom).}
    \label{loss_balls}
  \end{center}
  \vspace{-0.5cm}
\end{figure}

\noindent
We experimented with the x1 ball and x3 ball cases. In both cases, the dimension of $\mathbf{v}_t$ is $10000$, and we changed the dimension of $\mathbf{h}_t$ to $10,30,100$, and the number of CD iterations $K$ to $3,10,20$. Learning curves are shown in Figure \ref{loss_balls}. 
\begin{figure}[H]
\vspace{-1cm}
  \begin{center}
    \includegraphics[width=6cm]{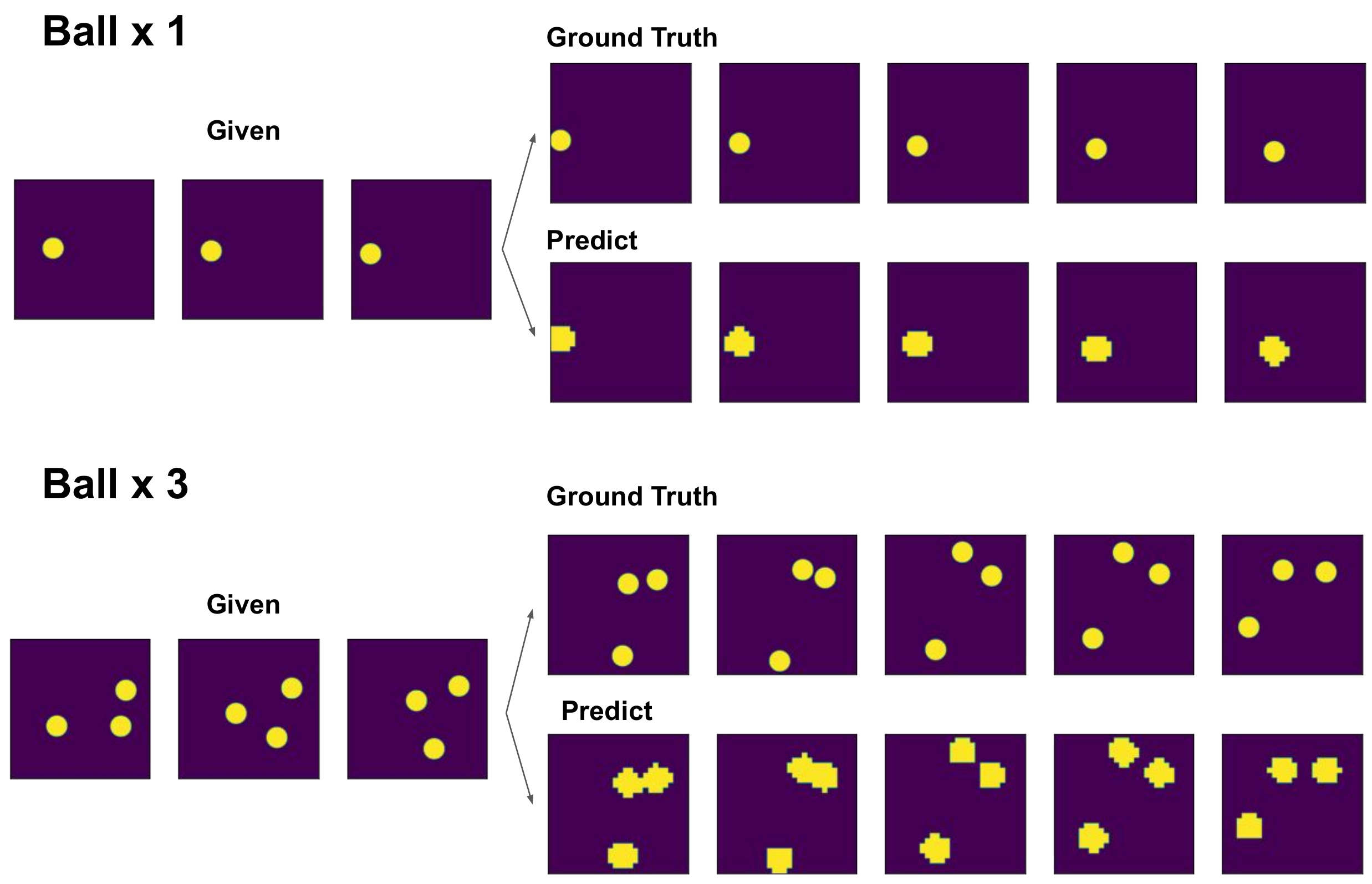}
    \vspace{-0.2cm}
    \caption{Ball state prediction ($\mathbf{h} = 100$). Three steps are given as input, and five steps are predicted. The top is ground truth, and the bottom is our prediction.}
  \label{ball_8steps}
  \end{center}
  \vspace{-1cm}
\end{figure}
\noindent
From Figure \ref{loss_balls}, x3 case is harder to learn the dynamics than x1 case. Loss is lower when the number of hidden variables is higher, and $K$ is fewer.
RTGB-RBM is trained rapidly in the first epoch, and learning progresses slowly from the second epoch.
The results show that the proposed method learns about bouncing balls early.
An example of the prediction by the trained RTGB-RBM is shown in Figure \ref{ball_8steps}.
In the x1 case, our model predicts \textit{bounce off} the wall.
In the x3 case, our model predicts that the balls will \textit{bounce off} each other.
This result shows that our model predicts not only the ball's trajectory but also the ball's bounce.

\subsection{Learned transition rules} 
\begin{figure}[H]
  \vspace{-0.5cm}
  \begin{center}
    \includegraphics[width=4.2cm]{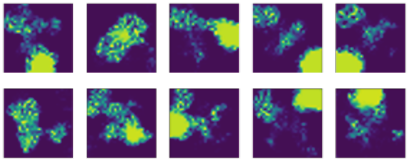}
      \vspace{-0.2cm}
    \caption{Feature map for x1 ball ($\mathbf{h}_t$ = $10$).}
    \label{feature_map_nh10}
  \end{center}
  \vspace{-1cm}
\end{figure}

\begin{figure}[H]
  \vspace{-1.6cm}
  \begin{center}
    \includegraphics[width=5cm]{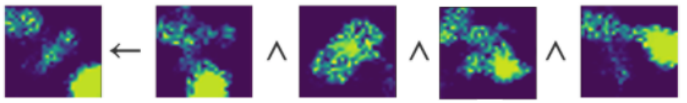}
  \end{center}
  \vspace{-0.6cm}
  \caption{Transition rule on feature maps corresponding to (\ref{learned_rule}) ($\mathbf{h} = 10$).}
  \vspace{-0.7cm}
  \label{transition_rule_high}
\end{figure}

\begin{figure}[H]
\vspace{-0.3cm}
  \begin{center}
    \includegraphics[scale=0.30]{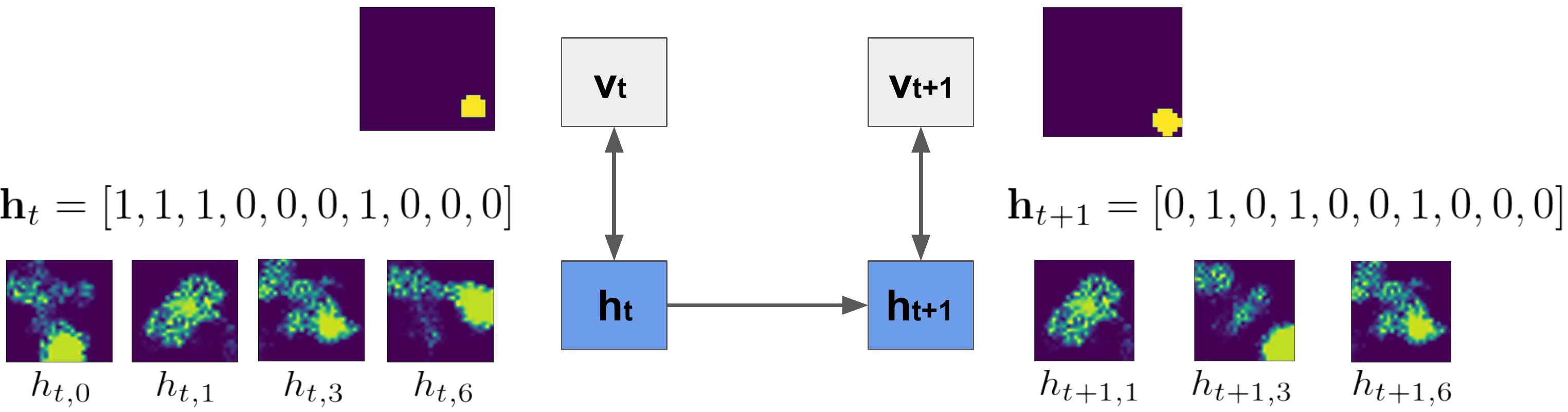}
  \end{center}
  \vspace{-0.6cm}
  \caption{Decode $\mathbf{v}_{t+1}$ from $\mathbf{h}_{t+1}$ by applying learned transition rules to $\mathbf{h}_t$.}
  \label{evaluation_transition_short}
\end{figure}

We describe the learned transition rules. 
To get a visual understanding of what the extracted hidden variables represent, we compute the feature map by applying the weight $\mathbf{W}$ to each hidden variable by (\ref{feature_map}).
\begin{eqnarray} \label{feature_map}
\vspace{-0.3cm}
v_{t,i} = \sigma(\sum_{j \in H} w_{ij} h_{t,j} + b_i) \ \ ( i \in V)
\vspace{-0.3cm}
\end{eqnarray}
The feature map of $\mathbf{v}$ for x1 ball case is calculated by (\ref{feature_map}) as shown in Figure \ref{feature_map_nh10}.
These feature maps imply the ball's position and direction. 
For example, the map in the top left corner represents that the ball is located near the center of the lower side.
The middle map above represents the ball moving from left to right. By combining these features, our model predicts the trajectory of the ball.
\noindent Among the learned state transition rules, the rule (\ref{learned_rule}) has the highest probability.
We evaluate the rule (\ref{learned_rule}) as an example.
Corresponding the learned rules with the feature maps, we get Figure \ref{transition_rule_high}.
The rule (\ref{learned_rule}) represents that if $h_{t,0}, h_{t,1}, h_{t,2},h_{t,6} = 1$, then $h_{t+1,3}$ become $1$ with probability $0.8732$.
Figure \ref{transition_rule_high} implies that $h_{t,0},h_{t,2}, h_{t,6}$ represent features that are trying to move in the lower right direction, and $h_{t,3}$ has a large value in the lower right corner. 
\begin{eqnarray}  \label{learned_rule}
0.8732 :: h_{t+1,3} \leftarrow h_{t,0} \wedge h_{t,1} \wedge h_{t,2}\wedge h_{t,6}
\end{eqnarray}
\vspace{-1cm}
\noindent
In Figure \ref{transition_rule_high}, the rule represents that the next feature map in the head is generated by combining the ball's current position and direction, represented by the four features in the body.
We extract such rules for all hidden variable state transitions. In Figure \ref{evaluation_transition_short}, we show an example of predicting $\mathbf{v}_{t+1}$ from $\mathbf{v}_t$ by using extracted rules. If we apply the learned rules to $\mathbf{h}_t=[1,1,1,0,0,0,1,0,0,0]$, we get $\mathbf{h}_{t+1}=[0,1,0,1,0,0,1,0,0,0]$. 
$\mathbf{v}_{t+1}$ is decoded from $\mathbf{h}_{t+1}$.
\vspace{-1cm}

\subsection{Comparative Experiment}
\begin{figure}[H]
  \vspace{-0.6cm}
  \begin{center}
    \includegraphics[scale=0.28]{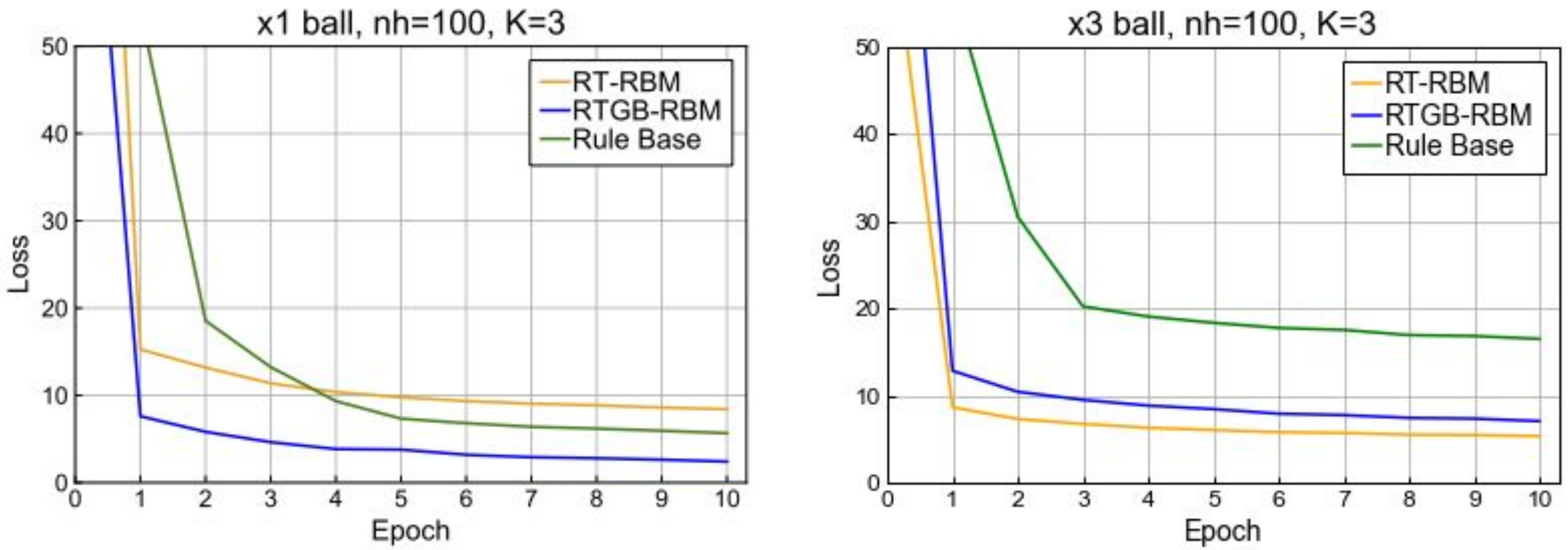}
  \end{center}
  \vspace{-0.6cm}
  \caption{Learning curve of x1 ball case (left) and x3 ball case (right) ($\mathbf{h} = 100$, $K=3$).}
  \label{loss_compare}
  \vspace{-0.6cm}  
\end{figure}

\noindent
We compare RT-RBM, RTGB-RBM, and rule-based predictions. Learning curves are shown in Figure \ref{loss_compare}. 
The results show that in x1 ball, both RTGB-RBM and rule-based predictions have higher accuracy than RT-RBM.
On the other hand, in x3 balls, RT-RBM performs better than the others, but RTGB-RBM performs as well as RT-RBM.

\begin{figure}[H]
  \vspace{-0.6cm}
  \hspace{-1.5cm}
  \begin{center}
    \includegraphics[scale=0.23]{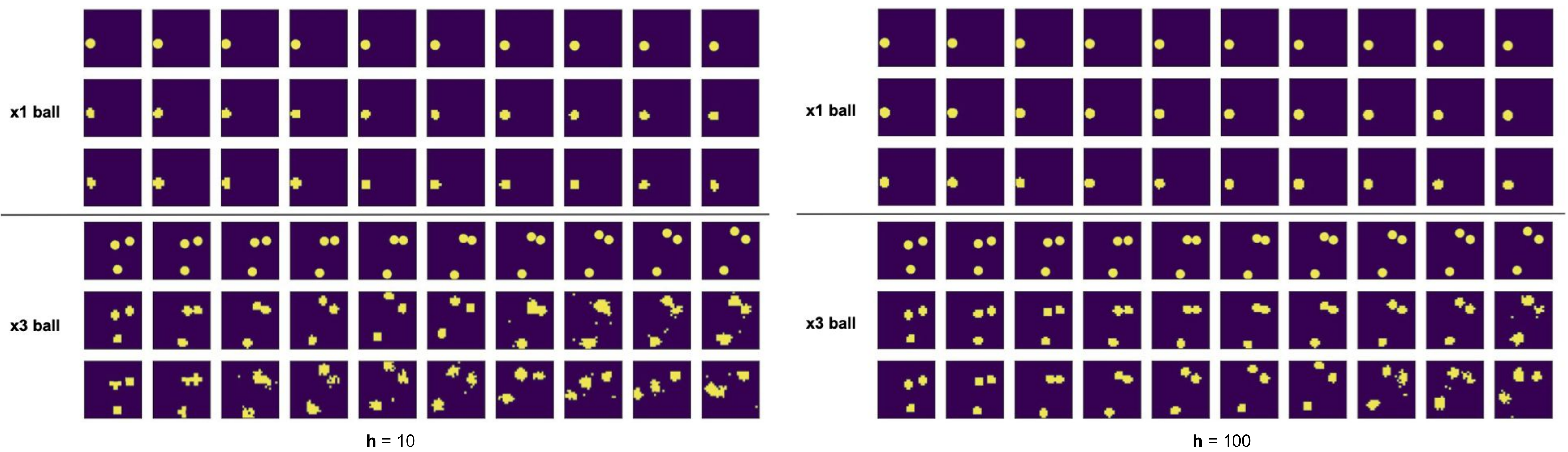}
  \end{center}
  \vspace{-0.6cm}
  \caption{Ground truth x1 and x3 ball trajectories (top), prediction of RTGB-RBM (middle), prediction using rules (bottom).  
  (Right: $\mathbf{h} = 10$, Left: $\mathbf{h} = 100$) }
  \label{compare_model_rule_wide}
  \vspace{-0.8cm}
\end{figure}

\noindent
Figure \ref{compare_model_rule_wide} (left) shows prediction of RTGB-RBM and prediction using rules, when $\mathbf{h} = 10$. For x1 ball, the trajectory of the ball is predicted by the rules and the RTGB-RBM. On the other hand, for x3 ball case predictions by RTGB-RBM are increasingly far from ground truth, and predictions by rules are no longer on the same trajectory. 
This result shows that rules between $10$ hidden variables were sufficient to predict the trajectory of the x1 ball, but they are not expressive enough to predict the trajectory of the x3 balls.

In Figure \ref{compare_model_rule_wide} (right), RTGB-RBM and rule-based predictions are improved, and they predict the ball’s trajectory and bounce.
This result indicates that the prediction accuracy increases with the number of hidden variables, even in the case of rule-based prediction.
As the number of hidden variables increases, state transitions can be expressed by rules in more detail. 
This indicates that some hidden variables are necessary to predict the dynamics.

\noindent
The rule-based method has lower prediction accuracy than the other two methods, and it predicts the ball's trajectory for the first five or six steps, gradually deviating from the ground truth. One possible reason is that since transitions based on rules occur with probability, noise and error increase as the state transitions forward, leading to incorrect predictions. 
Furthermore, it is difficult to describe the dynamics of multiple objects, such as three balls, in our rule form.
We guess the rules for each ball are not expressive enough, and we need to use rules that can represent interactions and relationships between balls. 
For example, the rules need to represent each ball's position and direction, and collisions.

Although there are still limitations as described above, our method can predict the state transition of the dynamics and learn rules between hidden variables which correspond to the features maps. Using hidden variables can reduce the size of state transition. While the original dynamics consist of many visible variables, this method can represent dynamics with few rules. Furthermore, since the rules are expressed in the form (\ref{rule_form}), they are interpretable, e.g., Figure \ref{transition_rule_high}, yet rule-based predictions are comparable to RTGB-RBM predictions, e.g., Figure \ref{loss_compare} (left).

\subsection{Moving MNIST}
We evaluate our method on Moving MNIST \cite{srivastava2015unsupervised}, which is more complex than the Bouncing Ball dataset.
Moving MNIST consists of videos of MNIST digits. It contains $10,000$ sequences; each sequence is $20$ frames long and consists of $2$ digits moving in a $64$x$64$ frame. Originally, each pixel takes a value of $0$ to $1$, but to outline the digit more distinct, we threshold each pixel value at $0.1$.
In Bouncing Ball, there were \textit{ball-to-ball} collisions and \textit{wall-to-ball} collisions, but in Moving MNIST, there are \textit{wall-to-digit} collisions but no \textit{digit-to-digit} collisions; the digits pass through each other. Therefore, we need to learn to reconstruct different digits, predict trajectories, bounce off walls, and predict after overlaps. 

\begin{figure}[H]
  \vspace{-0.5cm}
  \begin{center}
    \includegraphics[scale=0.38]{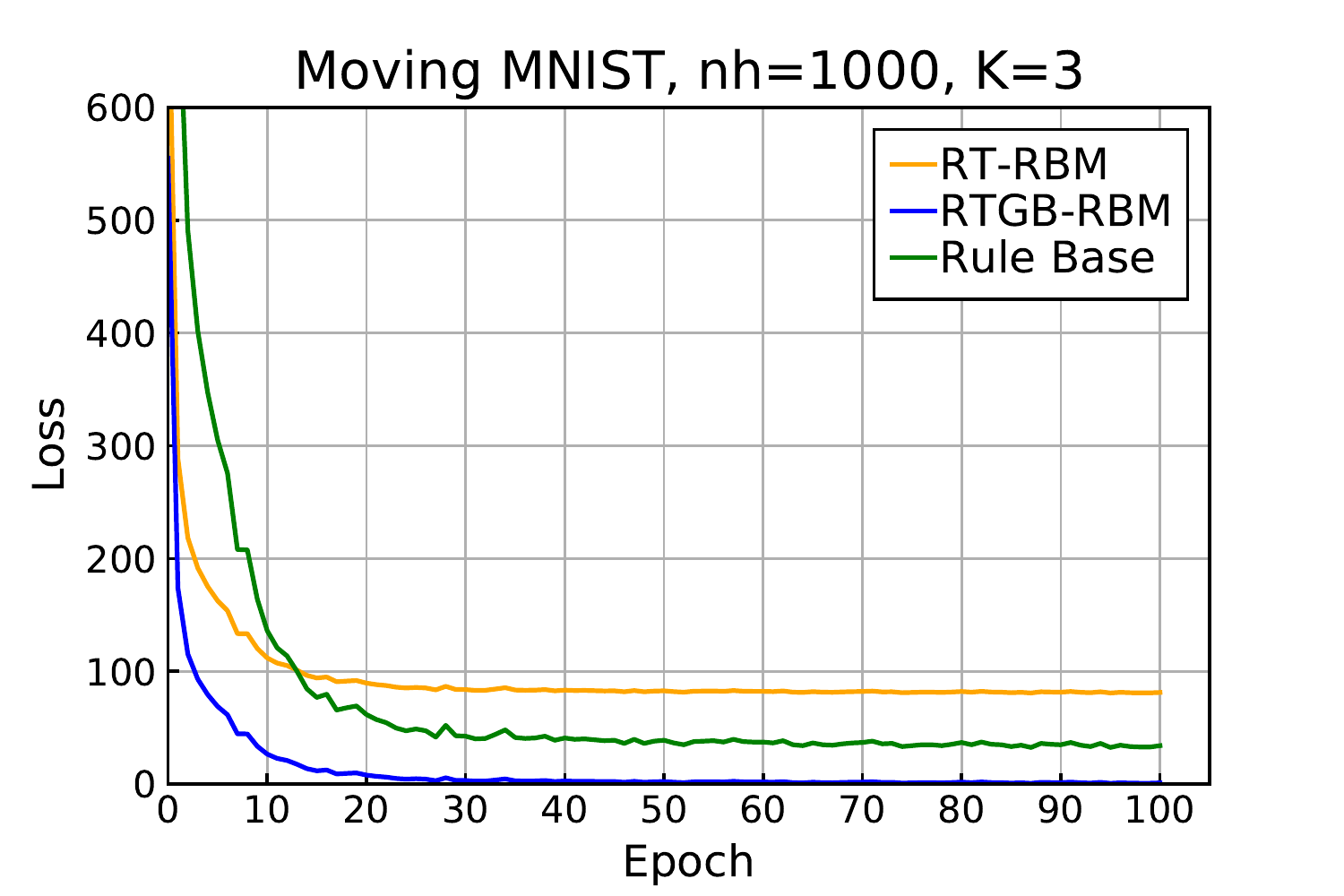}
  \end{center}
  \vspace{-0.8cm}
  \caption{Learning curve on moving MNIST ($\mathbf{h} = 1000$, $K=3$).}
  \label{loss_compare_movingmnist}
  \vspace{-0.6cm}  
\end{figure}

The learning curves for RT-RBM, RTGB-RBM, and rule-based on moving MNIST are illustrated in Figure \ref{loss_compare_movingmnist}.
The training is performed with the first $5$ frames as input and the remaining $15$ frames as predictions.
We train models $100$ epochs, and the loss in each epoch is calculated by (\ref{loss}).
The figure shows that RTGB-RBM prediction is better than RT-RBM.
As training progressed, the rule-based method approached the accuracy of RTRGB-RBM and eventually became more accurate than RT-RBM.

\begin{figure}[H]
  \vspace{-0.7cm}
  \begin{center}
    \includegraphics[scale=0.23]{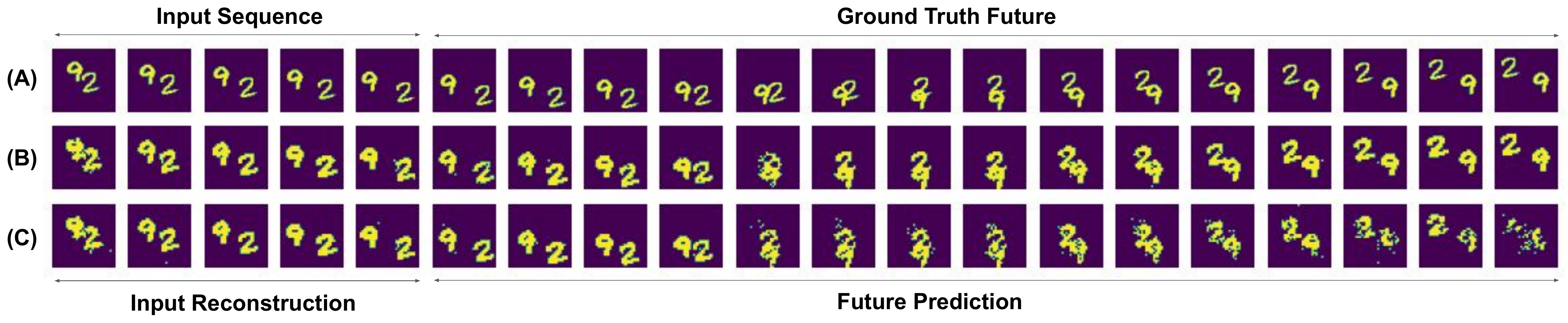}
  \end{center}
  \vspace{-0.6cm}
  \caption{
   Prediction of RTGB-RBM and Rule-based method when $h=1000$ and $K=3$: (A) Ground truth, (B) RTGB-RBM, (C) Rule-based.
   Input Reconstruction is decoded by giving the observed data to the visible layer as input, and Future Prediction is decoded only from the information in the hidden layer.}
  \label{prediction_compare_movingmnist_2}
\end{figure}

\noindent
An example of RTGB-RBM and rule-based prediction is shown in Figure \ref{prediction_compare_movingmnist_2}.
There are numbers $2$ and $9$. Our methods reconstruct the different shapes of the numbers and predict their trajectories.
When the digits overlap, the prediction becomes ambiguous, and it seems impossible to distinguish the digits from the ambiguous frame.
Nevertheless, it can predict the trajectory after overlap. From the result, we guess the hidden layers contain some features to distinguish the digits, and the transitions between hidden layers also contain enough information to reconstruct their trajectories after overlap. 
In addition, although the rule-based prediction is less accurate than the RTGB-RBM prediction, it can predict trajectories because the essential information is preserved in extracted rules.

\section{Conclusion} \label{sec5}
In this study, considering that real-world data has both discrete and continuous values and temporal relationships, we proposed RTGB-RBM, which combines GB-RBM to handle continuous visible variables and RT-RBM to capture time dependence between discrete hidden variables. We also proposed a rule-based method that extracts essential information as hidden variables and represents state transition rules in interpretable form. The experimental results show that our methods can predict future states of Bouncing Ball and Moving MNIST datasets. Furthermore, by corresponding the learned rules with the features represented by the hidden variables, we found that those rules contain essential information to determine $\mathbf{v}_{t+1}$ from the information at time $\mathbf{v}_t$. The rule-based method reduced observable data consisting of many variables into interpretable rules consisting of a few variables and performed well enough to predict the dynamics. However, more comprehensive experiments are needed to compare our method with other related methods on various dynamic datasets. Moreover, we should show through theoretical analysis why the proposed method works well. 
In future work, we will improve the rule-based method by learning rules with a rich structure that can handle multiple object interactions and longer dependencies. Furthermore, we aim to learn rules consisting of visible and hidden variables and rules consisting of continuous and discrete values.
\bibliography{reference}
\bibliographystyle{unsrt}
\end{document}